\documentclass{article}
\usepackage{spconf}
\usepackage{amsmath}
\usepackage{graphicx}
\usepackage{hyperref}

\title{DeGraF-Flow: Extending DeGraF Features for accurate and efficient sparse-to-dense optical flow estimation\vspace{-0.5cm}}
\name{ Felix Stephenson$^{1}$, Toby P. Breckon$^{1}$, Ioannis Katramados$^{2}$ \vspace{-0.5cm}}

\address{\vspace{-1in}Durham University, UK$^1$ $|$ NHL Stenden University of Applied Sciences, Netherlands$^2$}

\begin{document}
\maketitle
\begin{abstract}
\vspace{-0.2cm}
Modern optical flow methods make use of salient scene feature points detected and matched within the scene as a basis for sparse-to-dense optical flow estimation. Current feature detectors however either give sparse, non uniform point clouds (resulting in flow inaccuracies) or lack the efficiency for frame-rate real-time applications. In this work we use the novel Dense Gradient Based Features (DeGraF) as the input to a sparse-to-dense optical flow scheme. This consists of three stages: 1) efficient detection of uniformly distributed Dense Gradient Based Features (DeGraF) \cite{Katramados2016a}; 2) feature tracking via robust local optical flow \cite{RLOF2016}; and 3) edge preserving flow interpolation \cite{Revaud2015} to recover overall dense optical flow. The tunable density and uniformity of DeGraF features yield superior dense optical flow estimation compared to other popular feature detectors within this three stage pipeline. Furthermore, the comparable speed of feature detection also lends itself well to the aim of real-time optical flow recovery. Evaluation on established real-world benchmark datasets show test performance in an autonomous vehicle setting where DeGraF-Flow shows promising results in terms of accuracy with competitive computational efficiency among non-GPU based methods, including a marked increase in speed over the conceptually similar EpicFlow approach \cite{Revaud2015}.
\end{abstract}

\begin{keywords}
optical flow, Dense Gradient Based Features, DeGraF, automotive vision, feature points \end{keywords}

\section{Introduction}
\label{sec:intro}
\vspace{-0.2cm}
Optical flow estimation is the recovery of the motion fields between temporally adjacent images within a sequence. Since its conception over 35 years ago \cite{HS, Lucas1981} it remains an area of intense interest in computer vision in terms of accurate and efficient computation for numerous applications \cite{flowsurvey}.

Dense optical flow estimation aims to accurately recover per-pixel motion vectors from every pixel in a video frame to the corresponding locations in the subsequent (or previous) image frame in the sequence. 
These vector fields form the basis for applications such as scene segmentation \cite{Sevilla-Lara2016a}, object detection and tracking \cite{Kale2015}, structure from motion and visual odometry \cite{Aqel2016}. The development of autonomous vehicles has revealed the necessity for real-time scene understanding \cite{Geiger2012}. This has lead to increased pressure to improve both the quality and computational efficiency of dense optical flow estimation.

To date, recent progress in this area has been driven by the introduction of increasingly challenging benchmark datasets \cite{Baker2011, Butler2012a, Geiger2012, Menze2015} providing accurate ground truth optical flow for comparison. Approaches which are robust to scene discontinuities (occlusions, motion boundaries) and appearance changes (illumination, chromacity) have shown strong results under static scene conditions \cite{Baker2011}. However, the recent KITTI optical flow benchmarks \cite{Geiger2012, Menze2015}, which comprise video sequences of dynamic real-world urban driving scenarios, presents additional challenges. In particular, significant motion vectors between subsequent video frames due to vehicle velocity through the scene, present a key problem in accurate optical flow estimation \cite{Brox2011} for such large displacement vectors. 

\begin{figure}[t]

\begin{minipage}[b]{0.5\linewidth}
  \centering
  \centerline{\includegraphics[width=4.2cm]{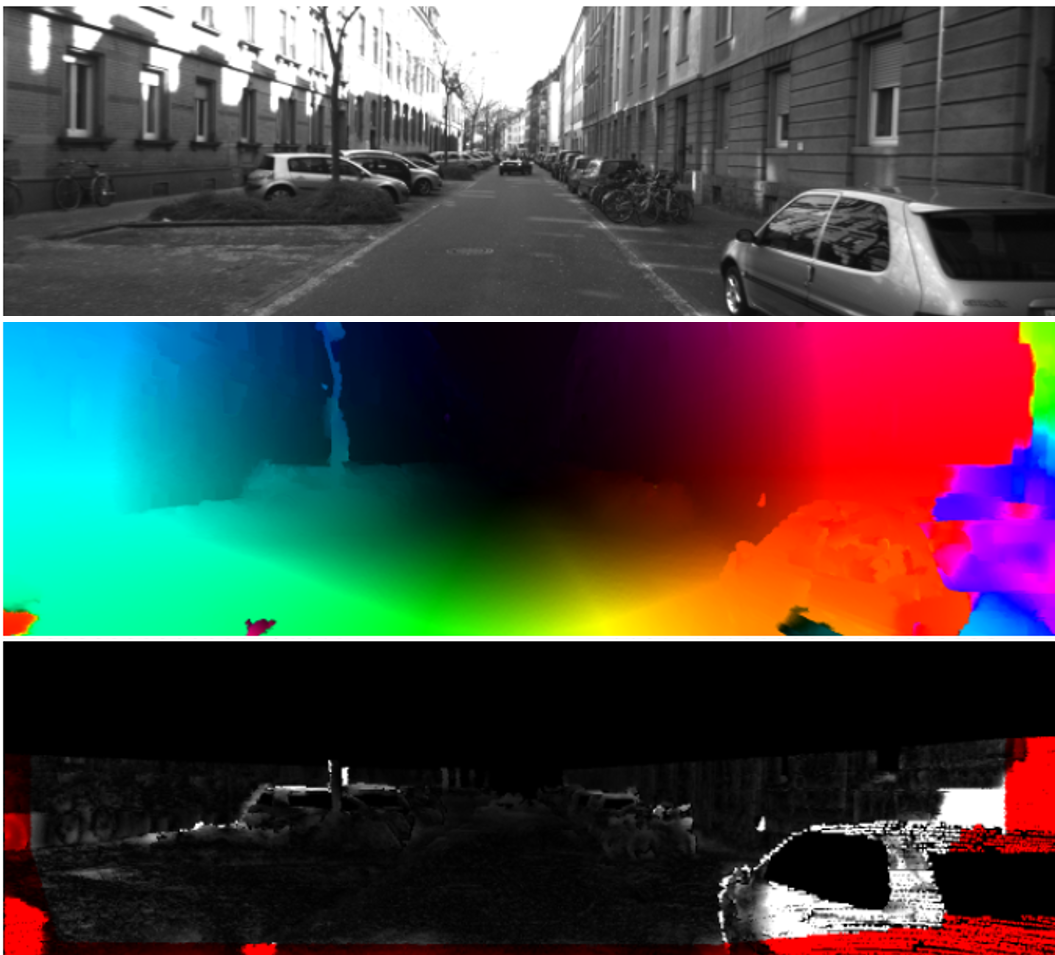}}
\end{minipage}
\begin{minipage}[b]{.48\linewidth}
  \centering
  \centerline{\includegraphics[width=4.2cm]{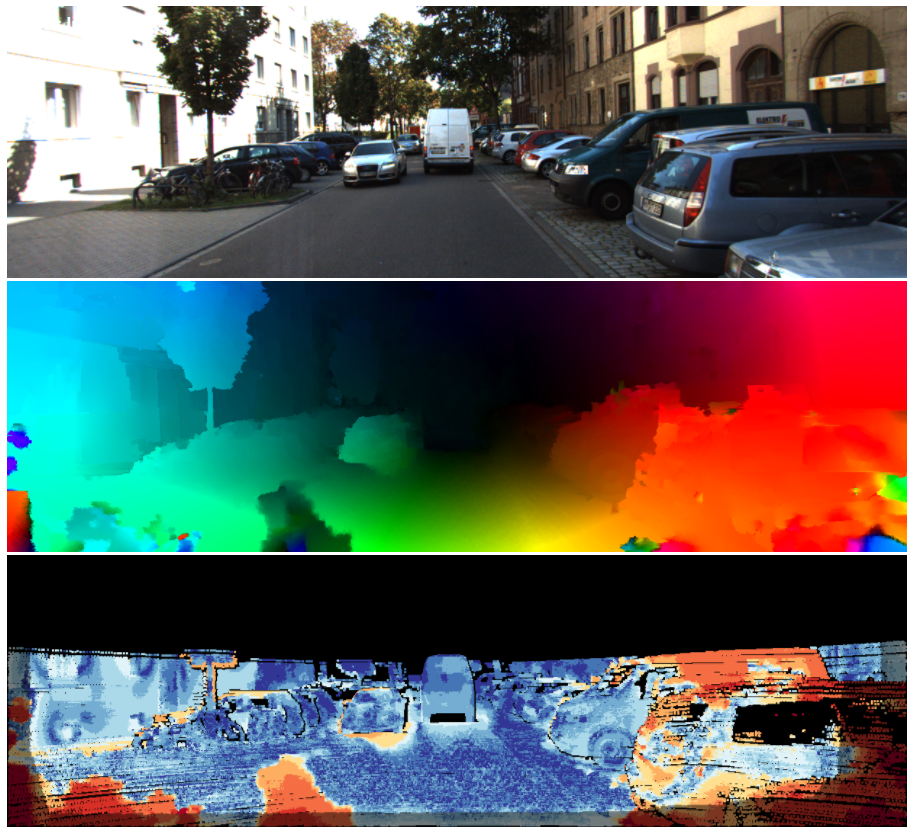}}
\end{minipage}

\caption{Dense optical flow results and error maps from KITTI 2012 \cite{Geiger2012} (left) and KITTI 2015 \cite{Menze2015} (right).}
\end{figure}

To cope with such challenges, contemporary optical flow methods use a sparse-to-dense estimation scheme, whereby a sparse set of points on a video frame are matched to points in the subsequent frame. The sparse set of optical flow vectors recovered from this matching are then used as input to a refinement scheme to recover dense optical flow \cite{Brox2011, briefmatch, Liu2015, Revaud2015, Timofte2015, Weinzaepfel2013}. Such matching of image points can accurately recover long range motions, however, as the matches are sparse (only cover a fraction of the image) they can lead to loss of accuracy at motion boundaries \cite{Revaud2015} in the refined dense optical flow. To address this issue, the recent work of \cite{Revaud2015} (EpicFlow) incorporates a novel state-of-the-art interpolator, Edge Preserving Interpolation of Correspondences (EPIC),  which recovers dense flow from sparse matches using edge detection to preserve accuracy at motion boundaries. This shows improved results compared against previous interpolation methods \cite{Wulff2015} and is currently used as a popular post-processing step in contemporary state-of-the-art dense optical flow estimation methods \cite{Bailer17, FullFlow, slowflow}.

Many of these sparse matching techniques are highly accurate but are notably incapable of real-time performance required for applications such as vehicle autonomy \cite{ Revaud2015, Timofte2015, Weinzaepfel2013, Yamaguchi2014}. 
By contrast, computational efficiency is more readily achievable via sparse point tracking whereby flow estimation only takes place on a fraction of the image. Interestingly, the seminal Lucas-Kanade \cite{Lucas1981} sparse point tracker proposed over 30 years ago still forms the basis for many contemporary state-of-the-art sparse flow techniques \cite{Bouguet2001, Garrigues2017, RLOF2016}. Robust Local Optical Flow (RLOF) \cite{RLOF2016} is one such derivative that shows state-of-the-art accuracy on the KITTI benchmark \cite{Geiger2012, Menze2015}. The most recent work on RLOF \cite{Geistert2017} demonstrates that combining sparse flow field with interpolators can achieve both efficient and accurate dense optical flow.

All methods that employ sparse feature tracking require a well defined feature set upon which matching can take place.  Furthermore, sparse flow vectors with uniform spatial coverage is an ideal for accurate dense optical flow recovery \cite{Geistert2017} making uniform feature distribution across the scene a key conduit to success. Common feature choices are Harris \cite{Harris1988} or FAST key-points \cite{Rosten2008} due to their relative speed. However these detectors do not guarantee uniform spatial feature distribution as they locate points only on highly textured image regions (e.g corners and edges). To address this issue, a current state of the art sparse flow method, denoted Fast Semi Dense Epipolar Flow (FSDEF) \cite{Garrigues2017}, forces uniformity of FAST key-point feature by use of a block-wise selection; this however adversely effects saliency causing increased erroneous matches. Further to this, FAST points do not provide sub-pixel precision so further refinement to the point matches is required to recover accurate optical flow. 
By contrast, recent work on the use of Dense Gradient Based Features (DeGraF) \cite{Katramados2016a} address these above issues. This approach provides spatially tunable feature density and uniformity in addition to sub-pixel accuracy. DeGraF has comparable speed to FAST and has been shown to be superior in terms of noise and illumination invariance \cite{Katramados2016a}. 

Motivated by these desirable attributes, our proposed method, DeGraF-Flow, takes a spatially uniform grid of DeGraF feature points as an input to a sparse-to-dense optical flow estimation scheme. Given two temporally adjacent images in a sequence, DeGraF points are detected in the first image and then efficiently tracked to the subsequent image using RLOF \cite{RLOF2016}. Finally dense optical flow is recovered using the established EPIC interpolation approach \cite{Revaud2015}.   

\begin{figure}[t]
  \centering
  \centerline{\includegraphics[width=8.5cm]{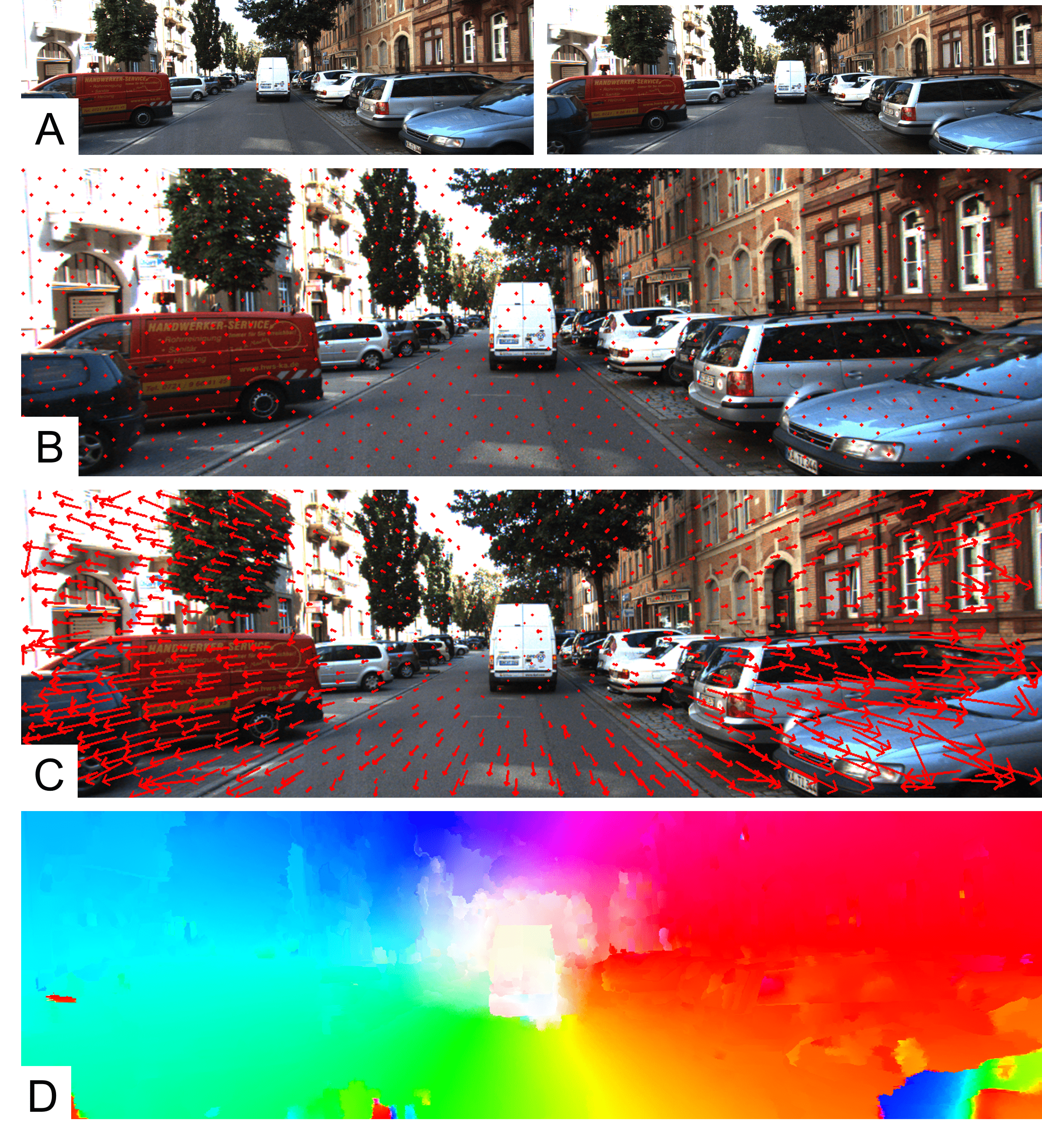}}
  \vspace{-0.5cm}
\caption{An overview of the DeGraF-Flow pipeline}
\end{figure}
\vspace{-0.3cm}
\section{Approach}
\label{sec:Approach}
\vspace{-0.3cm}
Dense optical flow is recovered from two temporally adjacent images (Figure 2A) using a three step process:

\textbf{Point detection} on the first image is carried out by calculation of an even grid of DeGraF points \cite{Katramados2016a} shown in Figure 2B. A sliding window is passed over the image with step size of $\delta$. A key-point is detected within the window at each step as follows.
  
For an image region $I$ of dimensions $w \times h$ containing grayscale pixels, two centroids, $C_{pos}$ and $C_{neg}$ are defined which define a gradient vector $\overrightarrow{C_{pos}C_{neg}}$ for the region. $C_{pos}$ is computed as the spatially weighted average pixel value:
\begin{equation}
\hspace{-0.18cm}
\resizebox{0.46\textwidth}{!}{$C_{pos}(x_{pos}, y_{pos}) = C_{pos}\left(\frac{\sum\limits_{i=0}^{h-1} \sum\limits_{j=0}^{w-1} i I(i,j)}{S_{pos}}, \frac{\sum\limits_{i=0}^{h-1} \sum\limits_{j=0}^{w-1} j I(i,j)}{S_{pos}}\right)$}
\end{equation}

where $S_{pos} = \sum\limits_{i=0}^{h-1} \sum\limits_{j=0}^{w-1} I(i,j)$. The negative centroid $C_{neg}$ is similarly defined as the weighted average of inverted pixel values: 
\begin{equation}
\resizebox{0.5\textwidth}{!}{$
C_{neg}(x_{neg}, y_{neg})= \\
C_{neg}\left(\frac{\sum\limits_{i=0}^{h-1} \sum\limits_{j=0}^{w-1} i(1+m- I(i,j))}{S_{neg}}, \frac{\sum\limits_{i=0}^{h-1} \sum\limits_{j=0}^{w-1} j (1+m-I(i,j))}{S_{neg}}\right)
$}
\end{equation}
where $S_{neg} = \sum\limits_{i=0}^{h-1} \sum\limits_{j=0}^{w-1}(1+ m - I(i,j))$ and $m = \text{max}_{(i,j)}I(i, j)$. Inverted pixel values are normalised ($1 \rightarrow 256$) to avoid division by zero.

 The key-point in each region is taken as the location of the most stable centroid, i.e if $S_{neg} > S_{pos}$ then the key-point is at $(x_{neg}, y_{neg})$ and vice versa. This choice is made because the larger value from $S_{neg}$ and $S_{pos}$ is less sensitive to noise and so the corresponding centroid is more robust. 

\textbf{Sparse Point tracking} of each DeGraF point to the subsequent image is carried out using the Robust Local Optical Flow approach of \cite{RLOF2016}. The sparse optical flow vectors recovered are shown in Figure 2C. 

\textbf{Interpolation} of sparse vectors to recover the dense flow field in Figure 2D is achieved using EPIC (Edge Preserving Interpolation of Correspondences) \cite{Revaud2015}. An affine transformation is fitted to the $k$ nearest support flow vectors, estimated using a geodesic distance which penalises crossing of image edges. This work uses image gradients instead of structured edge maps for defining image edges. The result is a dense optical flow field estimation as illustrated in Figure 2D. 

\vspace{-0.3cm}
\section{Evaluation}
\label{sec:Evaluation}
\vspace{-0.3cm}

Evaluation is carried out on the KITTI optic flow estimation benchmark data sets (denoted as KITTI 2012 \cite{Geiger2012} and KITTI 2015 \cite{Menze2015}).

Statistical accuracy is measured using the established End-Point Error (EPE) metric \cite{Geiger2012, Menze2015}. For a predicted optical flow vector $\mathbf{u}^p$ at every pixel with corresponding ground flow truth vector $\mathbf{u}^{gt}$, the EPE is then defined as the average difference between the predicted and ground truth vectors over the image:

\begin{equation}
\text{EPE} = \frac{1}{N} \sum_i \lVert \mathbf{u}_i^p - \mathbf{u}_i^{gt} \rVert ^2,
\end{equation}

where $N$ is the number of pixels and EPE is hence measured in pixels.

Our method (DeGraF / DeGraF-Flow) was implemented in C++ and all experiments run on a Core i7 using four CPU cores. All timings reported are for the run-time of the algorithm excluding image input/output and display.

\textbf{Parameters:} DeGraF window size, $w=h=3$ and step size, $\delta=9$.  For RLOF, the global motion and illumination models are used as per \cite{RLOF2016} with the adaptive cross based support region described in \cite{RLOF2014}. This algorithm is termed RLOF(IM-GM) in reported results in Table 2. For EPIC we use $k=128$  \cite{Revaud2015}.

\vspace{-0.3cm}
\subsection{Comparison of Feature Detectors}
\label{ssec:points}
\vspace{-0.1cm}

To justify the use of DeGraF points, we compare with established feature detectors for dense flow computation on KITTI 2012 \cite{Geiger2012}.

As is shown in \cite{Geistert2017}, when using RLOF and interpolation to recover dense optical flow, a uniform grid of points is a superior input compared to other feature point detectors. Here we repeat the experiment of \cite{Geistert2017} but with the addition of DeGraF (Table 1).

\begin{table}
\tiny
\centering
\resizebox{\columnwidth}{!}{%
\begin{tabular}{c | c | c | c}
Point Detector & \# Points & EPE & \begin{tabular}[c]{@{}c@{}}Detection\\ Time(s) \end{tabular} \\ \hline
DeGraF \cite{Katramados2016a} & 5400  & \textbf{1.34} & 0.07 \\
SURF \cite{Bay2006}           & 5282  & 1.43 & 0.90   \\
SIFT \cite{Lowe1999}          & 5400  & 1.77 & 1.80   \\
AGAST \cite{agast2010}        & 5624  & 1.92 & 0.11 \\
FAST \cite{Rosten2008}        & 5562  & 2.88 & 0.07 \\
ORB \cite{Rublee2011}         & 5400  & 5.60  & 0.28 \\
\end{tabular}
}
\caption{Comparison of differing point detectors for dense optical flow computation on KITTI 2012 example \#164 with each tuned to detect approximately 5400 points per image.\vspace{-0.3cm}}
\end{table}

Table 1 shows the EPE on a KITTI image pair from using popular feature detectors for the first stage of optical flow estimation. To allow meaningful comparison, each detector is tuned to ensure a comparable number of points are detected. DeGraF shows the best performing EPE and efficient detection, equal to FAST.
\vspace{-0.3cm}
\subsection{Benchmark Comparison - KITTI 2012}
\label{ssec:2012}
\vspace{-0.1cm}
The KITTI 2012 benchmark \cite{Geiger2012} comprises 194 training and 194 test image pairs (1240 $\times$ 376 pixels) depicting static road scenes.  Semi dense (50\%) ground truth, calculated using a LiDAR, is provided.

Table 2 shows the results on the KITTI test set for CPU methods that can process an image pair in under 20 seconds. All such methods that perform better than DeGraf-Flow in terms of EPE are included in the table. Not all less accurate methods are shown. Below the dividing line the best sparse flow methods are shown. Note how, although these methods have very low error, this error is only reported over a greatly reduced density of points. The result of standalone Pyramidal Lucas-Kanade (PLK) \cite{Bouguet2001} is shown as a baseline reference. The first two columns give the percentage of estimated flow vectors that have an EPE of more than 3px. The next two columns give average EPE values over the test set. Noc denotes statistics on only the non occluded pixels, where occluded pixels are those which appear in the first image but not in the second. Methods are ranked in order of increasing non-occluded outlier percentage (Out-Noc).

DeGraF-Flow shows promising results in terms of balancing computational efficiency (run-time, Table 2) and accuracy. In Table 2 it ranks thirteenth in accuracy (Out-Noc) and is shown to be the \textit{fourth fastest CPU method} that has a percentage outlier of non occluded pixels of less than 10\%. Other faster methods such as PCA-Flow and DIS-Fast show significantly higher EPE . PCA-Flow and PCA-Layers both employ an interpolation scheme for computing dense flow. The superior results of DeGraF-Flow show that the EPIC interpolator is the correct choice, which agrees with the findings in \cite{Geistert2017}. At the time of testing, DeGraF-Flow was the 18th fastest overall, including GPU methods with an overall accuracy rank of 56 from a total of 95 submissions within KITTI 2012.

EpicFlow and RLOF(IM-GM) are highlighted in Table 2 as these are the two constituent components of our method. DeGraF-Flow is outperformed by EpicFlow but runs almost five times faster. RLOF(IM-GM) represents near state of the art in sparse optical flow, making it an excellent candidate for tracking DeGraF points. RLOF is second only to FSDEF \cite{Garrigues2017} which is comtemporary work to that presented here.

\begin{table*}[t]
\tiny
\centering
\label{my-label}
\resizebox{\textwidth}{!}{%
\begin{tabular}{c | c | c | c | c | c | c | c}
{\bf Method} & {\bf Out-Noc} & {\bf Out-All} & {\bf Avg-Noc} & {\bf Avg-All} & {\bf Density} & {\bf Run-time} & {\bf Environment}\\ \hline
SPS-Fl \cite{Yamaguchi2014} & 3.38 \% & 10.06 \% & 0.9 px & 2.9 px & 100.00 \% & 11 s & 1 core @ 3.5 Ghz\\
SDF \cite{Jeevith2011} & 3.80 \% & 7.69 \% & 1.0 px & 2.3 px & 100.00 \% & TBA & 1 core @ 2.5 Ghz\\
MotionSLIC \cite{Yamaguchi2013} & 3.91 \% & 10.56 \% & 0.9 px & 2.7 px & 100.00 \% & 11 s & 1 core @ 3.0 Ghz \\
RicFlow \cite{Hu2017}& 4.96 \% & 13.04 \% & 1.3 px & 3.2 px & 100.00 \% & 5 s & 1 core @ 3.5 Ghz \\
CPM2 \cite{Li2017} & 5.60 \% & 13.52 \% & 1.3 px & 3.3 px & 100.00 \% & 4 s & 1 core @ 2.5 Ghz \\
CPM-Flow \cite{Li2017}& 5.79 \% & 13.70 \% & 1.3 px & 3.2 px & 100.00 \% & 4.2s & 1 core @ 3.5 Ghz \\
MEC-Flow \cite{mecflow} & 6.95 \% & 17.91 \% & 1.8 px & 6.0 px & 100.00 \% & 3 s & 1 core @ 2.5 Ghz \\
DeepFlow \cite{Weinzaepfel2013} & 7.22 \% & 17.79 \% & 1.5 px & 5.8 px & 100.00 \% & 17 s & 1 core @ 3.6Ghz \\
RecSPy+ \cite{Hu2018} & 7.51 \% & 15.96 \% & 1.6 px & 3.6 px & 100.00 \% & 0.16 s & 1 core @ 2.5 Ghz \\
RDENSE(anon) & 7.72 \% & 14.02 \% & 1.9 px & 4.6 px & 100.00 \% & 0.5 s & 4 cores @ 2.5 Ghz \\
\textbf{EpicFlow} \cite{Revaud2015} & 7.88 \% & 17.08 \% & 1.5 px & 3.8 px & 100.00 \% & 15 s & 1 core @ 3.6 Ghz \\
SparseFlow \cite{Timofte2015}  & 9.09 \% & 19.32 \% & 2.6 px & 7.6 px & 100.00 \% & 10 s & 1 core @ 3.5 Ghz \\
\textbf{DeGraF-Flow} & 9.41 \% & 16.93 \% & 2.5 px & 8.4 px & 100.00 \% & 3.2 s & 4 cores @ 2.5 Ghz\\
PCA-Layers \cite{Wulff2015} & 12.02 \% & 19.11 \% & 2.5 px & 5.2 px & 100.00 \% & 3.2 s & 1 core @ 2.5 Ghz \\
PCA-Flow \cite{Wulff2015} & 15.67 \% & 24.59 \% & 2.7 px & 6.2 px & 100.00 \% & 0.19 s & 1 core @ 2.5 Ghz \\
DB-TV-L1 \cite{Zach2007} & 30.87 \% & 39.25 \% & 7.9 px & 14.6 px & 100.00 \% & 16 s & 1 core @ 2.5 Ghz \\
DIS-FAST \cite{Kroeger2016} & 38.58 \% & 46.21 \% & 7.8 px & 14.4 px & 100.00 \% & 0.023s & 1 core @ 4 Ghz \\

\hline
FSDEF \cite{Garrigues2017}& 1.07 \% & 1.17 \% & 0.7 px & 0.7 px & 41.81 \% & 0.26s & 4 cores @ 3.5 Ghz \\
\textbf{RLOF(IM-GM)} \cite{RLOF2016} & 2.48 \% & 2.64 \% & 0.8 px & 1.0 px & 11.84 \% & 3.7 s & 4 core @ 3.4 Ghz \\
RLOF \cite{RLOF2012}& 3.14 \% & 3.39 \% & 1.0 px & 1.2 px & 14.76 \% & 0.488 s & GPU @ 700 Mhz\\
BERLOF \cite{RLOF2013} & 3.31 \% & 3.60 \% & 1.0 px & 1.2 px & 15.26 \% & 0.231 s & GPU @ 700 Mhz\\
PLK \cite{Bouguet2001} & 27.44 \% & 31.04 \% & 11.3 px & 17.3 px & 92.33 \% & 1.3 s & 4 cores @ 3.5 Ghz\\
\end{tabular}}
\vspace{-0.2cm}
\caption{KITTI 2012 Benchmark Results - comparison of the best CPU methods that can process an image pair in under 20 seconds. Below the line = sparse flow algorithms; Noc = pixels that are not occluded in the second image; First two columns = percentage of flow vectors that have an EPE of greater than 3px; methods are ordered by Out-Noc. Full table of results can be found on the KITTI benchmark website \cite{kitti}.\vspace{-0.2cm}}

\end{table*}

\begin{table*}[t]
\tiny
\centering
\resizebox{0.9\textwidth}{!}{%
\begin{tabular}{c | c | c | c | c | c | c}
{\bf Rank} & {\bf Method} & {\bf Fl-$bg$} & {\bf Fl-$fg$} & {\bf Fl-$all$} & {\bf Run-time} & {\bf Environment}\\ \hline
65 & EpicFlow \cite{Revaud2015}  & 25.81 \% & 28.69 \% & 26.29 \%  & 15 s & 1 core @ 3.6 Ghz \\
67 & DeepFlow \cite{Weinzaepfel2013} & 27.96 \% & 31.06 \% & 28.48 \%  & 17 s & 1 core @ 3.5 Ghz  \\
68 & DeGraF-Flow  & 28.78 \% & 29.69 \% & 28.94 \% & 3.2 s & 4 cores @ 2.5 Ghz \\
\end{tabular}
}
\vspace{-0.2cm}
\caption{KITTI 2015 Benchmark Results - percentage of pixels with EPE $\geq$ 3 pixels are given; $fg$ and $bg$ refer to motion of foreground moving vehicles and the static background scene respectively. \vspace{-0.2cm} }
\end{table*}

\vspace{-0.3cm}
\subsection{Benchmark Comparison - KITTI 2015}
\label{ssec:2015}
\vspace{-0.1cm}

The KITTI 2015 benchmark \cite{Menze2015} comprises 200 training and 200 test image pairs  (1242 $\times$ 375 pixels) with the increased challenges of dynamic scene objects (vehicles).  These exhibit far larger pixel displacements in some areas resulting in lower algorithm performance on KITTI 2015 (Table 3) than on KITTI 2012 (Table 2).

Table 3 shows the results of DeGraF-Flow on the test set compared to DeepFlow \cite{Weinzaepfel2013} and EpicFlow \cite{Revaud2015} (which both use DeepMatch as the sparse matching technique). The percentage of flow vectors with an EPE greater than 3px are shown, with $fg$ and $bg$ referring to foreground objects and the background scene respectively.

As with the KITTI 2012 benchmark results our approach shows comparable accuracy with significantly reduced run-time over EpicFlow and DeepFlow. Over all submissions to the KITTI 2015 benchmark, DeGraF-Flow reports \textit{the 10th fastest CPU method}. In terms of accuracy it places 13th (out of 22) from CPU methods that take less than than 10 seconds to process an image pair. Over the total of 90 submissions DeGraF-Flow places 68 in terms of Fl-all and 20 in terms of run-time.

\vspace{-0.2cm}
\section{Conclusion}
\vspace{-0.3cm}

In this paper we present DeGraF-Flow, a novel optical flow estimation method. DeGraF-Flow uses a rapidly computed grid of Dense Gradient based Features (DeGraF) and then combines an existing state of the art sparse point tracker (RLOF \cite{RLOF2016}) and interpolator (EPIC \cite{Revaud2015}) to recover dense flow. With only minimal impact on accuracy (within 2\% of EPIC-Flow \cite{Revaud2015} and DeepFlow \cite{Weinzaepfel2013} across all metrics) our approach offers significant gains in computational performance for dense optic flow estimation.

We show that the invariability, density and uniformity of DeGraF points yield superior dense flow results compared to other popular point detectors. The rapid end-to-end optic flow estimation time is also conducive to real-time applications such as scene understanding for future autonomous vehicle applications.

On the KITTI 2012 and 2015 benchmarks \cite{Geiger2012, Menze2015} our method shows competitive run-time and comparable accuracy with other CPU methods. Future work will exploit the tracking of DeGraF features for applications in an autonomous vehicle setting.

\bibliographystyle{IEEEbib}
\footnotesize\bibliography{CVproject}

\end{document}